%% file: main.tex
\RequirePackage{fix-cm}
\documentclass{fairmeta}
\microtypesetup{expansion=false}\usepackage{amsmath,amssymb}
\usepackage{longtable,colortbl}
\usepackage{wrapfig}
\usepackage{float}

\DeclareFontShape{T1}{optimistic}{b}{n}{<->ssub * optimistic/m/n}{}
\setcitestyle{numbers,square,sort&compress}
\usepackage{fontawesome5}
\newcommand{\ProjectURL}{https://yilinj10.github.io/RevolveVLN/}
\newcommand{\CodeURL}{https://github.com/YilinJ10/RevolveVLN}
\newcommand{\ModelURL}{https://huggingface.co/yilin-jia/RevolveVLN}
\newcommand{\ResourceLink}[2]{#2\ifdefempty{#1}{}{\enspace\href{#1}{\nolinkurl{#1}}}}
\renewcommand{\metadatalist}{%
  
  \begin{minipage}[b]{0.70\linewidth}
    \fontsize{10}{12}\selectfont\sffamily
    \ResourceLink{\ProjectURL}{\textcolor{metablue}{\faGlobe}\, \textbf{Project:}}\par
    \ResourceLink{\CodeURL}{\textcolor{metafg}{\faGithub}\, \textbf{Code:}}\par
    \ResourceLink{\ModelURL}{\textcolor{metablue}{\faCube}\, \textbf{Model:}}%
  
  \end{minipage}\par\vspace{0.05cm}\noindent\hfill
  \begin{minipage}[b]{0.34\linewidth}
    \raggedleft
    
    {
      \raisebox{-0.17cm}{\includegraphics[trim=0 0 1420bp 0,clip,height=0.72cm]{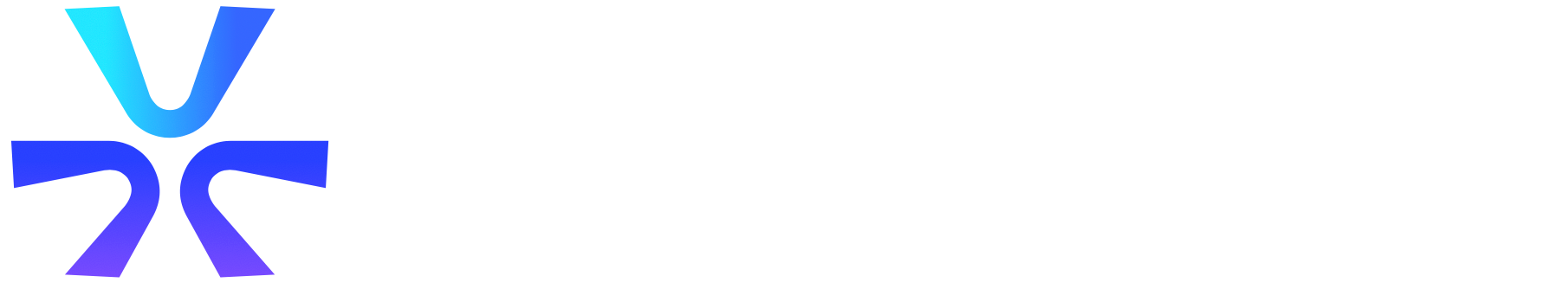}}%
      \hspace{0.10cm}{\fontsize{13}{15}\selectfont\sffamily\color{metafg}Beta Infinity}%
    }%
  \end{minipage}%
}

\title{RevolveVLN: Evolving Egocentric Memory towards Spatial Intelligence in Vision-Language Navigation}
\author[1,3,*]{Yilin Jia}
\author[2,*]{Yangyi Zhong}
\author[3]{Hongze Yao}
\author[3]{Mai Liu}
\author[3]{Xiaohui Lv}
\author[3]{Shuai Tao}
\author[3]{Wulong Liu}
\author[2]{Rong Li}
\author[2]{Zeying Gong}
\author[4]{Tianshuai Hu}
\author[1,\dagger]{Junfeng Fan}
\author[2,4,\dagger]{Junwei Liang}
\affiliation[1]{Institute of Automation, Chinese Academy of Sciences}
\affiliation[2]{The Hong Kong University of Science and Technology (Guangzhou)}
\affiliation[3]{Beta Infinity}
\affiliation[4]{The Hong Kong University of Science and Technology}
\contribution{\textsuperscript{*} Equal contribution\quad \textsuperscript{$\dagger$} Corresponding authors}
\abstract{Vision-Language Navigation (VLN) requires spatial memory that evolves with new observations and the agent’s changing pose. Geometry-enhanced VLN methods build spatial memory for vision-language models by enriching past frames with features from geometry foundation models. However, these frames remain tied to their original geometry and only implicitly related to the agent, limiting spatial intelligence. We observe that streaming geometry foundation models already offer an untapped opportunity to update them as new observations arrive. We propose RevolveVLN, a framework that turns this opportunity into \textit{Evolving Egocentric Memory}. At each step, it refreshes representations of past observations with the latest streaming geometry and relates them to the agent’s current pose through explicit geometric transformations. Trained only on R2R-CE and RxR-CE with monocular RGB input, RevolveVLN sets a new state of the art under the same training setting; when modestly scaled, it further matches large-scale navigation foundation models like LightNav-0 and Qwen-RobotNav trained on up to 30× more additional data. Furthermore, it achieves zero-shot gains over all models, including GPT-6.1-sol, on NavSpace, a benchmark for spatial intelligence.}
\begin{document}
\maketitle

\input{sections/01-introduction}

\input{sections/02-related-work}

\input{sections/03-method}
\input{sections/04-experiments}

\input{sections/05-conclusion}

\section*{LLM Usage Statement}
Generative AI tools were used to assist with literature search, code implementation, figure formatting, and manuscript editing. The authors take full responsibility for the experimental design, scientific claims, reported results, and final manuscript.

\clearpage
{
  \small
  \bibliographystyle{ieeenat_fullname}
  \bibliography{main}
}
\clearpage
\input{sections/appendix}

\end{document}

%% file: sections/01-introduction.tex
\section{Introduction}
\label{sec:introduction}

Vision-Language Navigation in Continuous Environments (VLN-CE) requires an agent to follow natural-language instructions through unseen scenes~\cite{krantz2020vlnce}. At every step, the agent decides where to go next from its current position and heading. Such decisions require a memory that records not only what the agent has seen, but also where each observation currently lies relative to the agent. For example, given the instruction “Return to the first doorway you passed.” the agent must know that the door is now behind it on the left, not straight ahead as it once was. Human navigation research calls this spatial updating~\cite{rieser1982spatial}: as people move, they continuously update the direction and distance of surrounding objects relative to themselves~\cite{wang2000updating}. In essence, navigation memory should evolve as the agent moves and remain anchored to its current pose.

Map-based methods once followed this principle explicitly. Egocentric maps are transformed by the agent’s motion and fused with new observations in its current frame~\cite{gupta2017cognitive,beeching2021egomap}. Topological graphs encode node directions and distances relative to the agent~\cite{chen2022duet,an2025etpnav}. Recently, pretrained vision-language models (VLMs) have brought stronger instruction and open-world visual understanding, shifting VLN toward VLM-based approaches. Some of these approaches still use depth and pose to build bird's-eye-view (BEV), semantic, or 3D maps~\cite{yang2026gavln,zhang2025mapnav,wang2025dynam3d}. Video-based methods instead feed frames directly into the VLM without mapping~\cite{zhang2024navid,zhang2025uninavid,cheng2025navila,wei2025streamvln}. Memory becomes a chronological sequence of snapshots, without explicitly reinterpreting the past with new observations or relating it to the current view. Spatial intelligence thus remains a weakness of these methods~\cite{yang2025thinking,wu2025spatialmllm,yang2026navspace}.

\newpage
\begin{wrapfigure}{r}{0.49\textwidth}
  \vspace{-\intextsep}
  \centering
     \includegraphics[width=\linewidth]{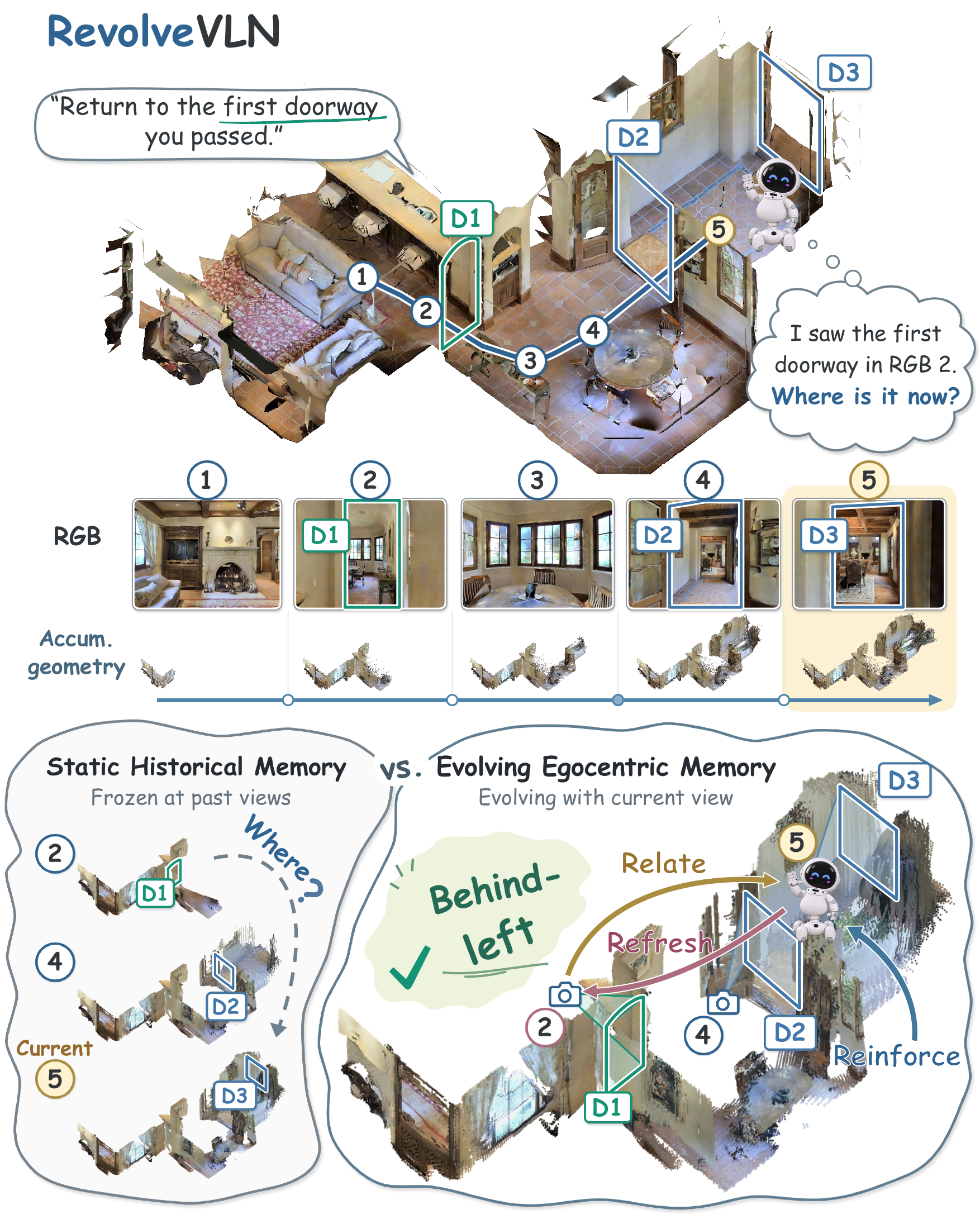}
  \caption{\textbf{Evolving Egocentric Memory in RevolveVLN.} At waypoint 5, following \emph{“Return to the first doorway you passed”}, the agent recognizes \texttt{D1} in RGB 2 but must locate it relative to its current view. Prior methods treat history as static records frozen at past views (\textbf{Static Historical Memory}), leaving \texttt{D1}'s current relative position unresolved (\texttt{"Where?"}). In contrast, \textbf{RevolveVLN} introduces \textbf{Evolving Egocentric Memory} that \textbf{Refreshes} past frames with accumulated geometry, \textbf{Reinforces} the current observation, and \textbf{Relates} history to the current view to locate \texttt{D1} to the \emph{behind-left}.}
  \label{fig:teaser}
\end{wrapfigure}

To address this weakness, navigation foundation models elicit spatial intelligence through large-scale spatial and embodied data~\cite{wang2026lightnav0,qwenrobotnav2026,chu2026abotn0}, but this requires tens of millions of training samples. Geometry-enhanced VLN methods instead use geometry foundation models~\cite{wang2025vggt,wang2025cut3r,wang2026pi3,zhuo2026streamvggt,lin2026depthanything3} to inject 3D structure into per-frame visual representations~\cite{zeng2025janusvln,liu2026dygeovln,fan2026prospect,zhang2026futurenav,pham2026adageovln,wang2026memvln}. Even so, historical frame representations remain fixed once formed. Their spatial relation to the present stays implicit in geometric features or absolute poses, so the VLM must infer it across viewpoints and over long horizons, precisely where VLMs struggle~\cite{deng2026lostinspace,lin2025ostbench,yang2026mmsibench}. Memory has gained geometry, but it still does not evolve with the agent.

However, we observe that the streaming geometry foundation models behind these methods already hold what their memory lacks~\cite{wang2025cut3r,zhuo2026streamvggt}: a current state that aggregates past observations, along with poses and 3D points in a shared coordinate frame~\cite{xu2026r3}. Geometry can thus drive memory to evolve with the agent from monocular RGB alone, without depth or mapping. We call this memory organization \textit{Evolving Egocentric Memory}: the content of each frame in memory is updated with the current state, and its position relative to the agent is recomputed from the current view. Such memory allows the agent to continually revise its understanding of the environment and navigate based on the spatial relations between past and present.

Accordingly, we propose RevolveVLN, a monocular VLN framework that realizes \textit{Evolving Egocentric Memory} through three sequential stages: Refresh, Reinforce, and Relate. All three stages share a single forward pass of the streaming geometry model. First, Refresh updates each frame in memory with the current state to reinterpret earlier observations. Next, Reinforce injects multi-layer geometry into the current view as a reference. Relate then queries memory from this view and aggregates past and current content via explicit relative geometry. 

Trained only on R2R-CE and RxR-CE with monocular RGB, RevolveVLN sets a new state of the art under the same training setting, outperforming several approaches with extra data or larger models~\cite{fan2026prospect,xin2026decovln,zeng2025janusvln}. When modestly scaled with 1.10M additional samples, it further reaches 69.3\% and 70.0\% success rates on R2R-CE and RxR-CE, matching or surpassing large-scale navigation foundation models trained on up to $30\times$ more data~\cite{wang2026lightnav0,qwenrobotnav2026}. On zero-shot NavSpace~\cite{yang2026navspace}, a benchmark for spatial intelligence in navigation, it outperforms all evaluated models, including GPT-6.1-sol. These results show that allowing memory to evolve with the agent effectively strengthens spatial intelligence in navigation.

In summary, our main contributions are threefold:
\begin{itemize}
  \item We formulate \textit{Evolving Egocentric Memory}, a memory paradigm that overcomes static VLM histories by continually updating past frames’ content and position relative to the agent.
  \item We propose RevolveVLN, a monocular framework that uses a single streaming geometry pass to Refresh past frames, Reinforce the current view, and Relate them via explicit relative geometry.
  \item We achieve state-of-the-art performance on R2R-CE and RxR-CE with strong data scalability, and outperform all evaluated models on zero-shot NavSpace.
\end{itemize}

%% file: sections/02-related-work.tex
\section{Related Work}
\label{sec:related_work}

\subsection{Vision-Language Navigation}

Vision-Language Navigation (VLN) requires an agent to follow language instructions to a target in discrete~\cite{anderson2018vln} or continuous~\cite{krantz2020vlnce} environments. Early methods store history chronologically via recurrent states or hierarchical encoding~\cite{hong2021vlnbert,chen2021hamt}. Subsequent work instead organizes spatial relations into explicit maps~\cite{wang2024hnr,wang2025g3dlf}, encoding relative node positions in topological graphs~\cite{chen2022duet,an2025etpnav} or aggregating multi-view features into BEV grids~\cite{an2023bevbert,wang2023gridmm} and 3D voxels~\cite{liu2024ver}. With the introduction of VLMs, map-based approaches continue using depth and pose to build BEV, semantic, or hierarchical 3D representations for decision-making~\cite{yang2026gavln,zhang2025mapnav,li2026hsgm,wang2025dynam3d,wang2026d3dvlp}. Video-based VLM navigators instead bypass map building, taking current and historical frames directly as context to predict actions~\cite{wang2026progressthink,guo2026awarevln,navforesee2026,song2025mgdm}. To control the growing history, they subsample or select historical frames~\cite{cheng2025navila,xin2026decovln}, compress visual tokens~\cite{zhang2024navid,zhang2025uninavid}, or apply sliding-context reuse and 3D redundancy removal~\cite{wei2025streamvln}. RevolveVLN likewise uses video frames as memory from monocular RGB alone; however, its memory evolves with the agent inside the VLM and remains referenced to the current view.

\subsection{Spatial Intelligence in VLMs and Navigation}

VLMs struggle with the metric and multi-view spatial relations implicit in RGB inputs~\cite{yang2025thinking,yang2026mmsibench}. Prior work either learns these relations from large-scale spatial instruction data~\cite{chen2024spatialvlm,cai2025spatialbot,song2025robospatial,bai2025qwen3vl,cai2026sensenovasi} or recovers 3D structure via geometry foundation models~\cite{wang2025vggt,wang2025cut3r,zhuo2026streamvggt,xu2026r3}, injecting geometry into visual tokens and layers~\cite{zheng2025vgllm,wu2025spatialmllm,fan2026vlm3r,zhang2026spatialstack,zhao2026spacemind,hu2026g2vlm,yu2026stream3d} or into attention~\cite{su2026dkvlm}. VLM navigators similarly split into data-driven foundation models trained on massive trajectories~\cite{zhang2026navfom,chu2026abotn0,wang2026lightnav0} and geometry-enhanced navigators that inject geometric features into per-frame tokens~\cite{zeng2025janusvln} or use geometry to refine injection, manage history, and predict future states~\cite{liu2026dygeovln,pham2026adageovln,fan2026prospect,zhang2026futurenav}. However, historical representations in these navigators remain static after arrival and lack explicit spatial relations to the current view, even with origin-referenced poses~\cite{wang2026memvln}. RevolveVLN instead reads the streaming geometry model's current state at each step, using poses and 3D points to explicitly compute view-relative spatial relations for memory.

\subsection{Geometry Foundation Models}

Geometry foundation models reformulate 3D reconstruction as feed-forward prediction, advancing from pairwise pointmap regression and matching~\cite{wang2024dust3r,leroy2024mast3r} to joint multi-view estimation of cameras, depth, and pointmaps~\cite{wang2025vggt,lin2026depthanything3,wang2026pi3}. They output explicit poses and 3D points while encoding cross-view structure in internal features. Yet joint multi-view processing requires recomputing the full sequence whenever a new frame arrives online. Streaming models instead process image streams incrementally to accumulate geometric context via persistent states and spatial memory~\cite{wang2025spann3r,wang2025cut3r,wu2025point3r} or causal attention over past views~\cite{zhuo2026streamvggt,lan2026stream3r,xu2026r3}. Among these, $R^3$ regresses pairwise relative poses and maintains a bounded keyframe bank for streaming inference. RevolveVLN builds on $R^3$, utilizing its accumulated states, poses, and 3D points.

%% file: sections/03-method.tex
\section{Method}
\label{sec:method}

\begin{figure}[!htbp]
  \centering
  \includegraphics[width=0.96\textwidth]{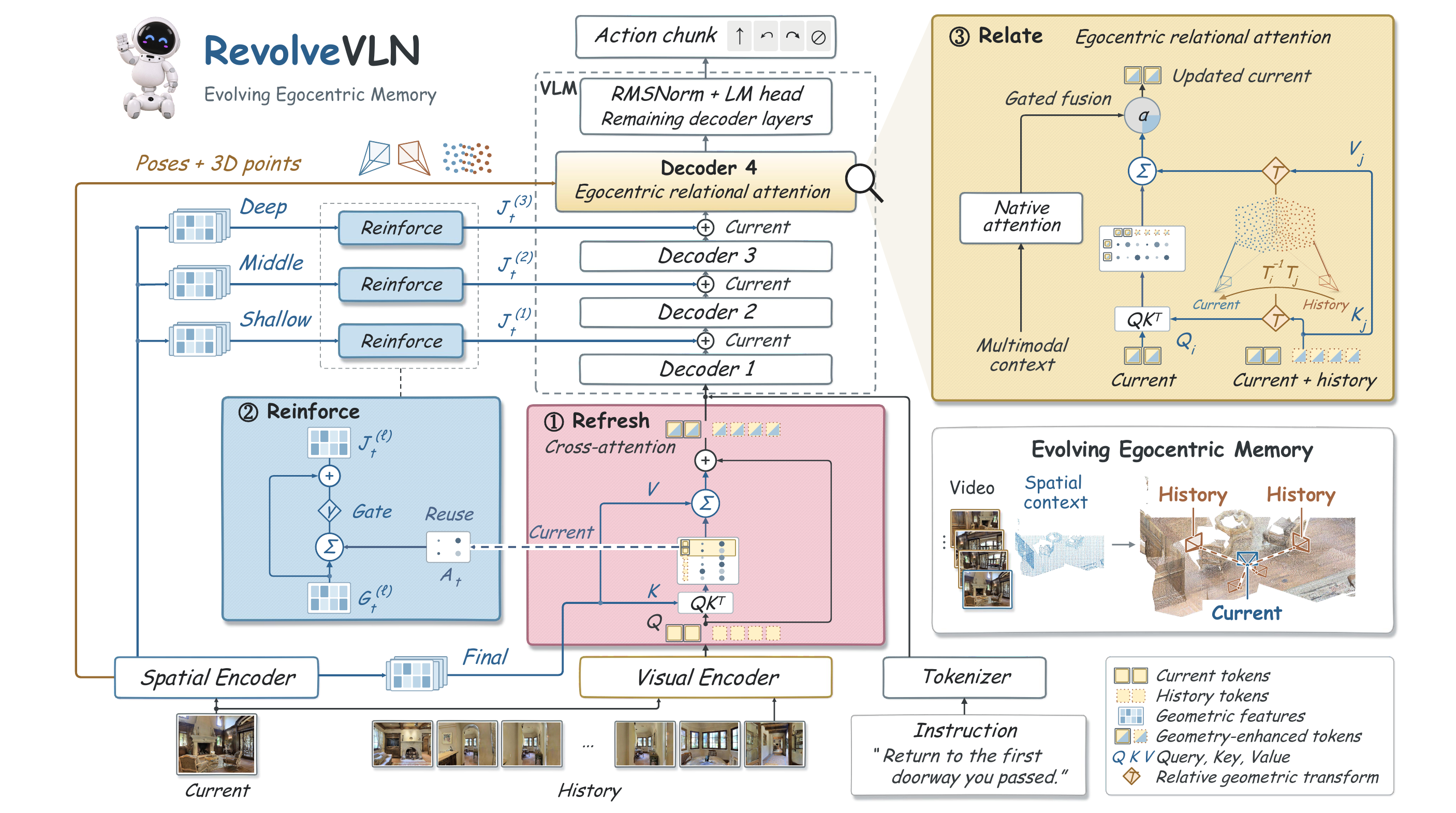}
  \caption{\textbf{Overview of RevolveVLN.} From monocular RGB, a streaming geometry foundation model (Spatial Encoder) extracts geometric features and estimates camera poses and 3D points. Combining video tokens with this spatial context, RevolveVLN builds  \emph{Evolving Egocentric Memory} to anchor historical observations to the current view through three stages: \textbf{(1) Refresh} updates current and historical visual tokens with accumulated geometry. \textbf{(2) Reinforce} reuses Refresh’s current-view attention weights for gated geometric enhancement across early decoder layers. \textbf{(3) Relate} transforms current and historical keys and values via relative geometry to update current tokens. The VLM combines the updated visual tokens with the instruction to predict an action chunk.}
  \label{fig:method_overview}
\end{figure}

Fig.~\ref{fig:method_overview} illustrates RevolveVLN. We first formulate the task in Sec.~\ref{sec:problem_formulation}, introduce our \textit{Evolving Egocentric Memory} in Sec.~\ref{sec:eem}, and detail its architecture in Sec.~\ref{sec:revolve_framework}.

\subsection{Problem formulation}
\label{sec:problem_formulation}

Given a natural-language instruction $x$, an agent navigates a continuous environment using monocular RGB observations $\mathcal O_{\le t}=(I_0,\ldots,I_t)$, where $I_t$ is the current view. At step $t$, the policy $\pi_\theta$ predicts a chunk of $L_t$ low-level actions:
\begin{equation}
\hat{\mathbf a}_t=\pi_\theta(x,\mathcal O_{\le t})\in\mathcal A^{L_t}, \qquad 1\le L_t\le4,
\label{eq:action_sequence}
\end{equation}
where $\mathcal A=\{\texttt{FORWARD},\texttt{LEFT},\texttt{RIGHT},\texttt{STOP}\}$, with $0.25\,\mathrm{m}$ forward steps and $15^\circ$ turns. The agent executes actions up to a configured horizon while receiving new observations, replanning iteratively until \texttt{STOP} is triggered.

\subsection{Evolving Egocentric Memory}
\label{sec:eem}

To keep historical observations useful for spatial decisions throughout long-horizon navigation, we introduce \textit{Evolving Egocentric Memory}. It structures navigation memory as content representations updated with accumulated geometric context and relative geometric relations updated with the agent's motion.

\paragraph{Streaming geometric context.}

We adopt the streaming feed-forward 3D reconstruction model R$^3$~\cite{xu2026r3} as our spatial geometry encoder, which retains cross-frame spatial context in a sparse keyframe bank and processes new frames with a bounded active cache. From a single forward pass on the current observation $I_t$, we extract implicit geometric features at zero-based layer indices 11, 15, and 19 (shallow, middle, and deep) along with final-layer features at index 23; after grid alignment and channel projection to match the visual tokens, these are denoted as $\{G_t^{(\ell)}\}_{\ell=1}^3$ and $G_t^{(*)}$, respectively. We also obtain explicit geometric cues—its predicted camera pose and back-projected 3D points corresponding to visual tokens.

\paragraph{From spatial encoding to memory evolution.}

Using this streaming geometric context, we construct and update each memory entry over time. At decision step $t$, the agent selects historical frame indices $\mathcal H_t\subseteq\{0,\ldots,t-1\}$ and combines them with the current frame to form the active view set $\mathcal F_t=\mathcal H_t\cup\{t\}$, from which a visual encoder extracts visual tokens $V_f$ ($f\in\mathcal F_t$). Each memory entry pairs a content representation $U_{f\mid t}$---obtained by enhancing $V_f$ with implicit geometric features and encoding it through the large language model (LLM)---with explicit relative geometry $\mathcal P_{f\to t}$, obtained by expressing the pose and 3D points of view $f$ in the current camera coordinate frame. Our \textit{Evolving Egocentric Memory} at step $t$ is thus formulated as:
\begin{equation}
\mathcal M_t=\{(U_{f\mid t},\mathcal P_{f\to t}):f\in\mathcal F_t\}.
\label{eq:eem}
\end{equation}
At each navigation step, $U_{f\mid t}$ updates with the accumulated spatial context while $\mathcal P_{f\to t}$ is recomputed from the current pose. This gives the same historical observation both an updated content interpretation and updated egocentric spatial relations. \textit{Evolving Egocentric Memory} thus provides a basis for cross-view association and spatial decision-making in long-horizon navigation within a limited visual-input budget.

\subsection{RevolveVLN framework}
\label{sec:revolve_framework}

To carry streaming spatial context through visual fusion and language decoding and realize \textit{Evolving Egocentric Memory}, we introduce a novel evolution mechanism that connects Refresh, Reinforce, and Relate in sequence.

\paragraph{Refresh: Updating visual memory with current geometric context.}
We design a memory refresh mechanism based on visual--geometric cross-attention. Positioned after the visual encoder and before visual tokens enter the LLM, it updates the selected views using the current streaming geometric context. The visual tokens $V_f$ of each view supply queries, while the current final-layer geometric features $G_t^{(*)}$ provide keys and values shared across views. Let $\phi_Q^h$, $\phi_K^h$, and $\phi_Z^h$ denote the learnable query, key, and value mappings for attention head $h$. The visual--geometric attention weights are
\begin{equation}
A_{f\leftarrow t}^{h}=\mathcal D\!\left[\operatorname{softmax}\!\left( \frac{\phi_Q^{h}(V_f)\,\phi_K^{h}(G_t^{(*)})^\top}{\sqrt{d_{\mathrm{ref}}}} \right)\right].
\label{eq:refresh_attention}
\end{equation}
Here, $d_{\mathrm{ref}}$ is the dimension of each head, and $\mathcal D$ denotes attention dropout. We use these weights to read geometric information and integrate it through a residual connection, yielding the refreshed visual tokens:
\begin{equation}
\widetilde V_{f\mid t}=\operatorname{FFN}_{\mathrm{res}}\!\bigl(V_f+\operatorname{Concat}_h\![A_{f\leftarrow t}^{h}\phi_Z^{h}(G_t^{(*)})]W_R\bigr).
\label{eq:refresh_tokens}
\end{equation}
Here, $\operatorname{Concat}_h$ concatenates head outputs along the channel dimension, $W_R$ projects the result back to the visual-token dimension, and $\operatorname{FFN}_{\mathrm{res}}$ denotes a feed-forward module with a residual connection. The same historical memory can thus continually update with spatial context. Reinforce further reuses the current view's attention weights $A_t^h\equiv A_{t\leftarrow t}^h$ for multilevel spatial enhancement.

\paragraph{Reinforce: Enhancing current understanding through shared correspondences.}
The current representation supplies queries, keys, and values for Relate. Its spatial understanding determines how memory is queried and how current content participates in aggregation. We adopt geometry injection into the first three LLM decoder layers~\cite{zhang2026spatialstack} and design a hierarchical enhancement mechanism based on visual--geometric correspondences, reusing Refresh's current-frame attention to enhance geometric features at different levels.

We partition the channels of $G_t^{(\ell)}$ according to the attention heads in Refresh and denote the $h$-th group by $G_t^{(\ell,h)}$. At each level $\ell\in\{1,2,3\}$, the enhanced geometric features $J_t^{(\ell)}$ combine geometry at corresponding positions with features read using $A_t^h$:
\begin{equation}
J_t^{(\ell)}=G_t^{(\ell)}+\tanh(\gamma_\ell)\operatorname{Concat}_h\![A_t^hG_t^{(\ell,h)}].
\label{eq:reinforce_features}
\end{equation}
Here, $\gamma_\ell$ is a learnable scalar for each level, initialized to zero. The direct term $G_t^{(\ell)}$ retains geometry at positions corresponding to the current visual tokens. The attention term adds spatial content from different levels through shared correspondences, avoiding the cost of recomputing query--key matching at each level.

At the output of the $\ell$-th LLM decoder layer, let $H_t^{(\ell)}$ denote the hidden states corresponding to the current frame's visual tokens. We inject the enhanced geometric features $J_t^{(\ell)}$ through a residual connection:
\begin{equation}
\widetilde H_t^{(\ell)}=H_t^{(\ell)}+J_t^{(\ell)}, \qquad\ell\in\{1,2,3\}.
\label{eq:reinforce_injection}
\end{equation}
The updated hidden states $\widetilde H_t^{(\ell)}$ proceed with the sequence to the next decoder layer. After the first three decoder layers, the hidden states of the selected views' visual tokens form $U_{f\mid t}$ and enter Relate in the fourth layer.

\paragraph{Relate: Organizing spatial interactions between past and present through relative geometry.}
We propose Egocentric Relational Attention, building on attention modulation through geometric transformations~\cite{miyato2024gta,su2026dkvlm}. It establishes the agent's current view as the spatial reference and directly embeds explicit spatial relations between past and present, updated as the agent moves, into attention matching and content aggregation. 

This mechanism is integrated into the fourth LLM decoder layer as a geometric side branch with $U_{f\mid t}$ as input. Let $\mathcal K_t$ be the set of visual-token indices across all selected views with valid poses and 3D points, and let $i$ index a current-frame visual token with a valid pose and 3D point. For a token $j\in\mathcal K_t$ from a selected view $f$, we construct an invertible block-diagonal geometric transformation $T_{j\mid t}$ from the relative geometry description $\mathcal P_{f\to t}$:
\begingroup
\thinmuskip=2mu\medmuskip=2mu\thickmuskip=3mu
\begin{equation}
T_{j\mid t}=\operatorname{blkdiag}\!\bigl(P_{j\mid t}^{(1)},\ldots,P_{j\mid t}^{(m)},C_{f\to t},\ldots,C_{f\to t},I\bigr),
\label{eq:geometric_transform}
\end{equation}
\endgroup
where
\begingroup
\thinmuskip=2mu\medmuskip=2mu\thickmuskip=3mu\setlength{\arraycolsep}{3pt}
\begin{equation}
P_{j\mid t}^{(r)}=\begin{bmatrix}R_{f\to t}&\lambda_r\mathbf p_{j\mid t}\\\mathbf0^\top&1\end{bmatrix},\quad C_{f\to t}=\begin{bmatrix}R_{f\to t}&\mathbf c_{f\to t}\\\mathbf0^\top&1\end{bmatrix}.
\label{eq:geometry_blocks}
\end{equation}
\endgroup
Here, $R_{f\to t}$ and $\mathbf c_{f\to t}$ are the rotation and translation of the source camera relative to the current camera, and $\mathbf p_{j\mid t}$ is the relative position of the token's corresponding 3D point in the current camera coordinate frame. The point geometry blocks $P_{j\mid t}^{(r)}$ encode token positions and source-view orientations using $m$ fixed scale coefficients $\lambda_r$, representing spatial positional relations between tokens. The camera geometry block $C_{f\to t}$ encodes relative poses between views. The two types of blocks act on different channel subspaces, with the identity transformation $I$ applied to the remaining channels. As the agent moves, $T_{j\mid t}$ is reconstructed with respect to the current view.

The $h$-th geometric attention head generates a query $q_i^h$ from the hidden state of current-frame token $i$, and keys $k_j^h$ and values $v_j^h$ from the hidden states of historical and current tokens $j\in\mathcal K_t$. The relative transformation $T_{i\mid t}^{-1}T_{j\mid t}$ modulates attention matching and value aggregation, yielding the geometric attention output $g_{i\mid t}^h$ for the current visual token:
\begingroup
\thinmuskip=2mu\medmuskip=2mu\thickmuskip=1mu
\begin{equation}
g_{i\mid t}^{h}=\sum_{j\in\mathcal K_t}\!\operatorname{softmax}_j\!\left(\frac{(q_i^h)^\top T_{i\mid t}^{-1}T_{j\mid t}k_j^h}{\sqrt{d_{\mathrm{rel}}}}\right)\!T_{i\mid t}^{-1}T_{j\mid t}v_j^h.
\label{eq:relate_attention}
\end{equation}
\endgroup
Here, $d_{\mathrm{rel}}$ is the dimension of each Relate head. The relative transformation acts on both keys and values.

Let $s_i$ denote the concatenated multi-head output for current visual token $i$ from the native attention, before output projection. We map both attention outputs to the LLM hidden space through the same learnable output projection $\phi_O$, initialized from the native attention's output projection and then trained independently. A learnable gate $\alpha=\sigma(\eta)$, with sigmoid $\sigma$ and learnable scalar $\eta$, combines them to produce the current visual token's attention output $o_i$:
\begin{equation}
o_i=(1-\alpha)\phi_O(s_i)+\alpha\phi_O\!\left( \operatorname{Concat}_h[g_{i\mid t}^h]\right).
\label{eq:relate_fusion}
\end{equation}
The fused attention output combines spatially enhanced visual content with explicit relative geometry to update only current-frame visual tokens with valid geometry. Other positions use the native attention output.

RevolveVLN thus continually updates historical and current content representations with streaming spatial context and organizes their interactions through relative geometry that changes as the agent moves. This realizes the joint evolution of memory content and egocentric relations in \textit{Evolving Egocentric Memory}, enhancing the agent's capacity for long-horizon spatial navigation.

%% file: sections/04-experiments.tex
\section{Experiments}
\label{sec:experiments}
\subsection{Experimental Setup}
\noindent\textbf{Datasets and metrics.} We evaluate on the validation-unseen splits of two VLN-CE benchmarks (R2R-CE and RxR-CE~\cite{krantz2020vlnce,ku2020rxr}) and assess zero-shot spatial generalization on NavSpace~\cite{yang2026navspace}, a benchmark for evaluating spatial intelligence in embodied navigation. For VLN-CE, we report navigation error (NE), oracle success rate (OSR), success rate (SR), success weighted by path length (SPL), and normalized dynamic time warping (nDTW). On NavSpace, we report per-category SR and macro-averaged NE, OSR, and SR across six spatial task categories: vertical perception (VP), precise movement (PM), viewpoint shifting (VS), spatial relationship (SpR), environment state (ES), and space structure (SS).

\noindent\textbf{Implementation details.} RevolveVLN pairs Qwen3-VL-4B-Instruct~\cite{bai2025qwen3vl} with the $R^3$ streaming geometry model~\cite{xu2026r3}, taking monocular RGB observations to predict a chunk of four actions and execute the first three at each replanning step. Throughout training, we keep $R^3$ and the visual encoder frozen and optimize the remaining parameters with AdamW at a global batch size of 512 across eight NVIDIA RTX PRO 6000 GPUs, using a 3\% linear warmup and cosine decay from a peak learning rate of $2 \times 10^{-5}$ for the language backbone. We first train for one epoch on R2R-CE and RxR-CE expert demonstrations; under the extra-data setting, we continue training with 1.10M additional samples by mixing original demonstrations, ScaleVLN~\cite{wang2023scalevln}, and DAgger~\cite{ross2011reduction} data at an approximate 40:40:20 ratio.

\subsection{Navigation Performance}
\input{tables/vln-ce-results}

\noindent\textbf{Navigation performance on VLN-CE.} Trained solely on benchmark demonstrations without extra data, $\text{RevolveVLN}^{\dagger}$ sets a new state of the art across all metrics under the same training regime in Table~\ref{tab:baseline-selected}. Specifically, it surpasses concurrent 4B geometry- and memory-based baselines---AdaGeoVLN~\cite{pham2026adageovln} by 8.5 and 11.2~pts in SR, and MemVLN~\cite{wang2026memvln} by 9.2 and 9.9~pts---while lowering NE to 4.3~m and 4.6~m. Remarkably, using only a 4B model and monocular RGB, $\text{RevolveVLN}^{\dagger}$ already outperforms heavier 7B/8B models with over 10M extra samples (e.g., NaVILA~\cite{cheng2025navila}, StreamVLN~\cite{wei2025streamvln}, JanusVLN~\cite{zeng2025janusvln}) and methods using panoramic and depth sensors (e.g., HNR~\cite{wang2024hnr}, GA-VLN~\cite{yang2026gavln}).

When modestly scaled with 1.10M additional samples, RevolveVLN achieves further substantial gains. On R2R-CE, it attains the highest SR (69.3\%) and OSR (74.4\%) while reducing NE to 3.5~m. On RxR-CE, it reaches 70.0\% SR, matches the best NE of 3.7~m, and achieves a dominant 72.9\% nDTW---exceeding Qwen-RobotNav~\cite{qwenrobotnav2026} (13.20M samples) by 4.3~pts and LightNav-0~\cite{wang2026lightnav0} (33.80M samples) by 5.5~pts. Matching or surpassing models trained on up to $30\times$ more data---particularly in NE and nDTW---underscores the strong geometric inductive bias of our \textit{Evolving Egocentric Memory}.

\input{tables/navspace-results}

\noindent\textbf{Zero-shot spatial generalization on NavSpace.} As shown in Table~\ref{tab:navspace}, RevolveVLN achieves the best zero-shot performance across all macro-averaged metrics without any NavSpace fine-tuning (33.7\% Macro SR, 46.7\% Macro OSR, and 4.1~m Macro NE), more than doubling general-purpose MLLMs (e.g., GPT-5) and substantially surpassing specialized navigation models (e.g., StreamVLN~\cite{wei2025streamvln} and the NavSpace-trained SNav~\cite{yang2026navspace}). Furthermore, our 4B model surpasses the proprietary reasoning upper bound GPT-6.1-sol (high) across all macro-averaged metrics (33.7\% vs. 30.8\% SR, 46.7\% vs. 36.7\% OSR, and 4.1 vs. 4.2~m NE) as well as five of the six spatial categories (VP, VS, SpR, ES, and SS), confirming that our \textit{Evolving Egocentric Memory} effectively structures streaming geometric cues into transferable 3D spatial representations. The remaining gap concentrates on PM (12.9\% vs. 43.8\%), which requires explicit metric step-counting (e.g., moving specific distances or angles) largely absent from landmark-driven VLN-CE training data.

\subsection{Ablation Studies}
\label{sec:ablation}

\begin{table}[!htbp]
\centering
\caption{\textbf{Component ablation on R2R-CE.}}
\label{tab:ablation-components}
\begingroup
\footnotesize
\setlength{\tabcolsep}{3pt}
\begin{tabular*}{\columnwidth}{@{}l@{\extracolsep{\fill}}rrrr@{}}
\toprule
Variant & NE$\downarrow$ & OSR$\uparrow$ & SR$\uparrow$ & SPL$\uparrow$ \\
\midrule
\textbf{Full} & \textbf{4.25} & \textbf{67.86} & \textbf{64.22} & \textbf{59.35} \\
w/o Refresh & 4.51 & 67.48 & 62.81 & 57.67 \\
w/o Reinforce & 5.19 & 51.22 & 42.79 & 38.23 \\
w/o Relate & 4.41 & 67.16 & 63.08 & 57.94 \\
\bottomrule
\end{tabular*}
\endgroup
\end{table}

\noindent\textbf{Component Ablation.} Table~\ref{tab:ablation-components} evaluates each module in RevolveVLN. Removing \textbf{Reinforce} causes the sharpest drop (SR from 64.22 to 42.79), as it serves as the essential structural bridge that associates and consolidates evolving memory across frames for the VLM. Removing \textbf{Refresh} lowers SR by 1.41~pts and SPL by 1.68~pts (NE 4.25m $\to$ 4.51m); notably, OSR stays comparable (67.48 vs. 67.86), showing that stale geometry mainly impairs precise stopping and path efficiency near the target. Excluding \textbf{Relate} reduces SR by 1.14~pts and SPL by 1.41~pts, confirming the benefit of explicit egocentric coordinate alignment. Together, they form an interdependent pipeline—supplying updated geometry (Refresh), linking it across space and time (Reinforce), and anchoring it to the current pose (Relate).

\begin{table}[!htbp]
\centering
\caption{\textbf{Design choice ablation on R2R-CE.}}
\label{tab:ablation-design}
\begingroup
\footnotesize
\setlength{\tabcolsep}{3pt}
\begin{tabular*}{\columnwidth}{@{}l@{\extracolsep{\fill}}rrrr@{}}
\toprule
Variant & NE$\downarrow$ & OSR$\uparrow$ & SR$\uparrow$ & SPL$\uparrow$ \\
\midrule
\textbf{Full} & \textbf{4.25} & \textbf{67.86} & \textbf{64.22} & \textbf{59.35} \\
Refresh: per-frame geometry & 4.41 & 66.88 & 62.26 & 57.57 \\
Reinforce: independent attention & 4.32 & 66.78 & 62.91 & 57.96 \\
Relate: multi-layer placement & 4.62 & 65.47 & 60.20 & 55.38 \\
\bottomrule
\end{tabular*}
\endgroup
\end{table}

\noindent\textbf{Streaming vs. Per-Frame Geometry (Refresh).} Replacing streaming geometry with each frame's isolated initial estimate (\emph{per-frame geometry}) drops SR by 1.96~pts (62.26) and SPL by 1.78~pts, performing even worse than omitting Refresh. This confirms that unaligned historical geometry introduces cross-step inconsistency.

\noindent\textbf{Shared vs. Independent Attention (Reinforce).} Computing separate attention across layers (\emph{independent attention}) instead of reusing Refresh's geometric attention (\emph{shared attention}) reduces SR by 1.31~pts and SPL by 1.39~pts, showing that shared geometric guidance aligns the current observation more reliably while keeping historical memory intact.

\noindent\textbf{Single- vs. Multi-Layer Placement (Relate).} Injecting explicit geometric transformations across multiple layers (\emph{multi-layer placement}, L4/L13/L19/L28/L34) instead of a single early layer (\emph{single-layer placement}, L4) drops SR by 4.02~pts and SPL by 3.97~pts, as single-layer injection at L4 sets a clean spatial anchor while deeper transforms disrupt high-level semantic reasoning.

\subsection{Qualitative Analysis}
\begin{wrapfigure}{r}{0.46\textwidth}
\centering
\includegraphics[width=\linewidth]{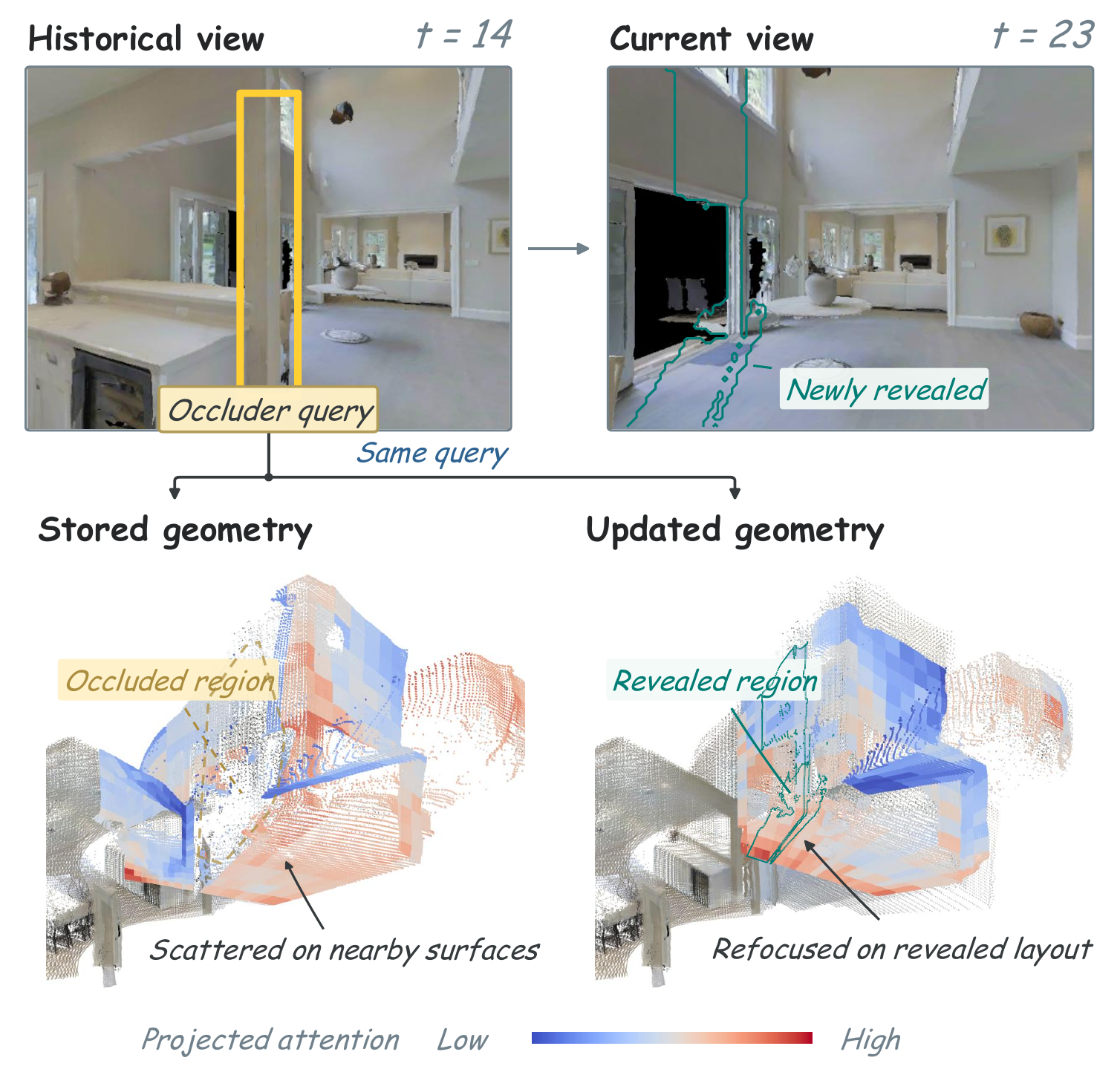}
\caption{\textbf{Refreshing historical observations with updated geometry.} The yellow box marks the historical occluder query at $t=14$ attending to stored ($t=14$) versus updated ($t=23$) geometry. As the agent moves forward and resolves the occlusion (green contour), cross-attention shifts from broad, uncertain surfaces onto newly revealed structures.}
\label{fig:history_reading}
\end{wrapfigure}
Due to space constraints, we focus on \textbf{Refresh} here and defer analyses of \textbf{Reinforce} and \textbf{Relate} to the Supplementary Material.

\noindent\textbf{Refresh.} To verify that \textit{Evolving Egocentric Memory} updates stale representations rather than keeping them static, Figure~\ref{fig:history_reading} probes the cross-attention of a historical region query when reading geometry before versus after occlusion clearance. Under early partial observation, incomplete 3D reconstruction yields high geometric uncertainty, causing the query to attend diffusely across nearby surfaces. Once new viewpoints resolve the missing layout, querying the refreshed geometry enables the same query to re-anchor onto newly exposed structural boundaries. This confirms that Refresh mitigates early observation bias without re-encoding raw frames.

\subsection{Real-World Deployment}
\label{sec:real-world}

\noindent\textbf{Deployment setup.}
We deploy RevolveVLN zero-shot on a commercial wheeled mobile robot without real-world fine-tuning, streaming forward-facing monocular RGB observations to a remote workstation (NVIDIA RTX 5090 D) for policy inference. Onboard, a local executor converts predicted four-action chunks into trajectories tracked by Nav2~\cite{macenski2020marathon2}, overlapping inference with robot motion.

\begin{figure}[!htbp]
  \centering
  \includegraphics[width=\textwidth]{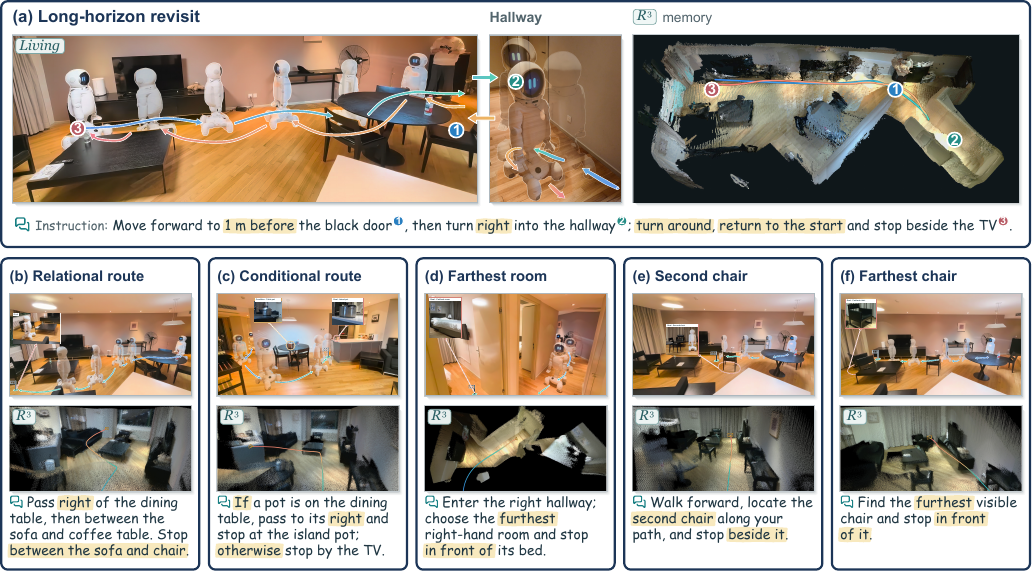}
  \caption{\textbf{Real-world navigation across six tasks.}
  Composited robot observations are paired with corresponding
  $R^3$ spatial memory. Trajectories are smoothed for visualization,
  and highlighted phrases indicate spatial instructions.}
  \label{fig:real_robot}
\end{figure}

\noindent\textbf{Real-world navigation analysis.}
We evaluate RevolveVLN across six zero-shot indoor tasks (Fig.~\ref{fig:real_robot}). In a challenging long-horizon revisit (Fig.~\ref{fig:real_robot}a), the policy navigates from the living room into the hallway ($1\rightarrow2$) and backtracks to stop beside the TV (3). Despite severe viewpoint shifts during $180^\circ$ turns, it preserves a drift-free global layout from monocular RGB alone, ensuring accurate closed-loop return. Across spatial reasoning (Fig.~\ref{fig:real_robot}b--f), RevolveVLN grounds language in its \textit{Evolving Egocentric Memory} to thread narrow relational gaps (between the sofa and coffee table, b), adapt trajectories to observed scene states (branching on a dining-table pot, c), and disambiguate relative spatial scales---including ordinal counts (\textit{second} chair, e) and metric distances (\textit{furthest} room/chair, d, f).

%% file: tables/vln-ce-results.tex
\begin{table}[!htbp]
\centering
\caption{\textbf{Navigation performance on VLN-CE validation-unseen splits.}}
\label{tab:baseline-selected}
\begingroup
\fontsize{8}{9}\selectfont
\setlength{\tabcolsep}{1.3pt}
\renewcommand{\arraystretch}{1.12}
\begin{tabular*}{\textwidth}{@{\extracolsep{\fill}}l*{14}{c}@{}}
\toprule
\multirow{2}{*}{Method} & \multirow{2}{*}{\shortstack{VLM\\params}} & \multicolumn{4}{c}{Observation} & \multicolumn{4}{c}{R2R-CE val-unseen} & \multicolumn{4}{c}{RxR-CE val-unseen} & \multirow{2}{*}{\shortstack{Extra\\data (M)}} \\
\cmidrule(lr){3-6}\cmidrule(lr){7-10}\cmidrule(lr){11-14}
 & & S.RGB & Pano. & Depth & Odom. & NE$\downarrow$ & OSR$\uparrow$ & SR$\uparrow$ & SPL$\uparrow$ & NE$\downarrow$ & SR$\uparrow$ & SPL$\uparrow$ & nDTW$\uparrow$ & \\
\midrule
ETPNav~\cite{an2025etpnav}\,{\tiny TPAMI'25} & -- &  & $\checkmark$ & $\checkmark$ & $\checkmark$ & 4.7 & 65.0 & 57.0 & 49.0 & 5.6 & 54.8 & 44.9 & 61.9 & -- \\
HNR~\cite{wang2024hnr}\,{\tiny CVPR'24} & -- &  & $\checkmark$ & $\checkmark$ & $\checkmark$ & 4.4 & 67.0 & 61.0 & 51.0 & 5.5 & 56.4 & 46.7 & 63.6 & -- \\
NaVid~\cite{zhang2024navid}\,{\tiny RSS'24} & 7B & $\checkmark$ &  &  &  & 5.7 & 49.2 & 41.9 & 36.5 & 5.7 & 45.7 & 38.2 & -- & 0.95 \\
NaVILA~\cite{cheng2025navila}\,{\tiny RSS'25} & 8B & $\checkmark$ &  &  &  & 5.2 & 62.5 & 54.0 & 49.0 & 6.8 & 49.3 & 44.0 & 58.8 & 13.13 \\
Uni-NaVid~\cite{zhang2025uninavid}\,{\tiny RSS'25} & 7B & $\checkmark$ &  &  &  & 5.0 & 57.4 & 51.8 & 47.7 & 5.7 & 56.1 & 44.5 & -- & 3.58 \\
StreamVLN~\cite{wei2025streamvln}\,{\tiny ICRA'26} & 7B & $\checkmark$ &  &  &  & 4.9 & 63.6 & 56.4 & 50.2 & 5.7 & 54.4 & 45.4 & 63.7 & 26.33 \\
JanusVLN~\cite{zeng2025janusvln}\,{\tiny ICLR'26} & 7B & $\checkmark$ &  &  &  & 4.8 & 65.2 & 60.5 & 56.8 & 6.1 & 56.2 & 47.5 & 62.1 & 10.69 \\
Progress-Think~\cite{wang2026progressthink}\,{\tiny CVPR'26} & 4B & $\checkmark$ &  &  &  & 4.7 & 63.6 & 60.1 & 53.6 & -- & -- & -- & -- & $\checkmark$ \\
D3D-VLP~\cite{wang2026d3dvlp}\,{\tiny CVPR'26} & 2B & $\checkmark$ &  & $\checkmark$ & $\checkmark$ & 4.7 & 67.2 & 61.3 & 56.1 & -- & -- & -- & -- & $\checkmark$ \\
PROSPECT~\cite{fan2026prospect}\,{\tiny RA-L'26} & 7B & $\checkmark$ &  &  &  & 4.9 & 65.2 & 58.9 & 54.0 & 5.7 & 54.6 & 46.2 & 62.1 & $\checkmark$ \\
DecoVLN~\cite{xin2026decovln}\,{\tiny CVPR'26} & 7B & $\checkmark$ &  &  &  & 5.0 & 63.5 & 56.3 & 50.5 & 5.7 & 54.2 & 46.3 & 63.5 & $\checkmark$ \\
GA-VLN~\cite{yang2026gavln}\,{\tiny CVPR'26} & 7B & $\checkmark$ &  & $\checkmark$ & $\checkmark$ & 4.8 & 67.6 & 61.0 & 55.2 & 5.9 & 55.4 & 45.2 & -- & $\checkmark$ \\
AdaGeoVLN~\cite{pham2026adageovln}\,{\tiny arXiv'26} & 4B & $\checkmark$ &  &  &  & 5.3 & 60.7 & 55.7 & 51.4 & 5.6 & 54.1 & 44.7 & 61.8 & 0 \\
GeoScaffold~\cite{geoscaffold2026}\,{\tiny arXiv'26} & 4B & $\checkmark$ &  &  &  & 5.1 & 61.9 & 55.3 & 51.0 & 5.8 & 53.2 & 46.2 & 59.8 & 0 \\
MemVLN~\cite{wang2026memvln}\,{\tiny arXiv'26} & 4B & $\checkmark$ &  &  &  & 5.2 & 61.2 & 55.0 & 50.5 & 5.6 & 55.4 & 48.4 & 64.6 & 0 \\
\midrule
{\fontsize{8.5}{9.5}\selectfont \textbf{RevolveVLN (Ours)}}$^\dagger$ & {\fontsize{8.5}{9.5}\selectfont 4B} & $\checkmark$ &  &  &  & {\fontsize{8.5}{9.5}\selectfont \textbf{4.3}} & {\fontsize{8.5}{9.5}\selectfont \textbf{67.9}} & {\fontsize{8.5}{9.5}\selectfont \textbf{64.2}} & {\fontsize{8.5}{9.5}\selectfont \textbf{59.4}} & {\fontsize{8.5}{9.5}\selectfont \textbf{4.6}} & {\fontsize{8.5}{9.5}\selectfont \textbf{65.3}} & {\fontsize{8.5}{9.5}\selectfont \textbf{55.4}} & {\fontsize{8.5}{9.5}\selectfont \textbf{68.4}} & {\fontsize{8.5}{9.5}\selectfont 0} \\
\midrule
Qwen-RobotNav~\cite{qwenrobotnav2026}\,{\tiny arXiv'26} & 4B & $\checkmark$ &  &  &  & 4.2 & 73.6 & 66.9 & 60.5 & 4.2 & 71.3 & 61.5 & 68.6 & 13.20 \\
LightNav-0~\cite{wang2026lightnav0}\,{\tiny arXiv'26} & 4B & $\checkmark$ &  &  &  & 3.9 & 73.7 & 68.5 & \textbf{62.8} & 3.7 & \textbf{73.6} & \textbf{64.5} & 67.4 & 33.80 \\
\midrule
{\fontsize{8.5}{9.5}\selectfont \textbf{RevolveVLN (Ours)}} & {\fontsize{8.5}{9.5}\selectfont 4B} & $\checkmark$ &  &  &  & {\fontsize{8.5}{9.5}\selectfont \textbf{3.5}} & {\fontsize{8.5}{9.5}\selectfont \textbf{74.4}} & {\fontsize{8.5}{9.5}\selectfont \textbf{69.3}} & {\fontsize{8.5}{9.5}\selectfont 61.3} & {\fontsize{8.5}{9.5}\selectfont \textbf{3.7}} & {\fontsize{8.5}{9.5}\selectfont 70.0} & {\fontsize{8.5}{9.5}\selectfont 59.2} & {\fontsize{8.5}{9.5}\selectfont \textbf{72.9}} & {\fontsize{8.5}{9.5}\selectfont 1.10} \\
\bottomrule
\end{tabular*}
\par\vspace{3pt}
\begin{minipage}{\textwidth}\fontsize{8}{9}\selectfont\raggedright
\textbf{S.RGB}, \textbf{Pano.}, \textbf{Depth}, and \textbf{Odom.} denote single-view RGB, panoramic RGB, depth, and odometry inputs. \textbf{Extra data (M)} lists additional training samples in millions (\textbf{0}: benchmark data only; $\checkmark$: undisclosed amount). $\dagger$: trained without additional navigation data, compared only with the baselines above it. Best results are in \textbf{bold}.\end{minipage}
\endgroup
\end{table}

%% file: tables/navspace-results.tex
\begin{table}[!htbp]
\centering
\caption{\textbf{Zero-shot spatial generalization on NavSpace.}}
\label{tab:navspace}
\begingroup
\fontsize{8}{9}\selectfont
\setlength{\tabcolsep}{0.9pt}
\renewcommand{\arraystretch}{1.08}
\begin{tabular*}{\columnwidth}{@{\extracolsep{\fill}}l*{6}{c}!{\hspace{2pt}}*{3}{c}@{}}
\toprule
& \multicolumn{6}{c}{Subtask SR (\%)} & \multicolumn{3}{c}{Macro average} \\
\cmidrule(lr){2-7}\cmidrule(l){8-10}
Method & VP & PM & VS & SpR & ES & SS & NE$\downarrow$ & OSR$\uparrow$ & SR$\uparrow$ \\
\midrule
LLaVA-Video-7B & 7.7 & 3.5 & 6.8 & 11.3 & 6.5 & 4.5 & 6.1 & 15.3 & 6.7 \\
Qwen2.5-VL-7B & 6.3 & 2.5 & 7.7 & 9.4 & 8.5 & 10.5 & 5.7 & 15.7 & 7.5 \\
GLM-4.5V-106B & 7.7 & 2.5 & 7.2 & 9.4 & 12.0 & 6.5 & 5.9 & 19.9 & 7.6 \\
\midrule
GPT-5 & 18.3 & 3.0 & 12.6 & 17.5 & 17.5 & 16.5 & 5.0 & 20.2 & 14.2 \\
Gemini-2.5-Pro & 23.6 & 4.0 & 9.2 & 7.1 & 13.0 & 10.0 & 5.0 & 17.2 & 11.2 \\
\midrule
NaVid & 23.1 & 7.0 & 22.7 & 24.1 & 33.0 & 10.0 & 4.9 & 30.7 & 20.0 \\
NaVILA & 3.4 & 2.5 & 5.3 & 3.8 & 8.0 & 13.0 & 6.3 & 8.8 & 6.0 \\
StreamVLN & 23.1 & 8.0 & 21.3 & 24.5 & 28.0 & 10.0 & 5.1 & 30.9 & 19.2 \\
SNav & 28.8 & 12.4 & 23.7 & 32.5 & 41.5 & 17.0 & 4.5 & 36.7 & 26.0 \\
\midrule
RevolveVLN (Ours)$^\dagger$ & 36.1 & 10.9 & 31.9 & 42.0 & 39.0 & 14.5 & 4.2 & 42.7 & 29.1 \\
\midrule
GPT-6.1-sol (high) & 24.0 & \textbf{43.8} & 32.4 & 28.3 & 36.5 & 20.0 & 4.2 & 36.7 & 30.8 \\
\midrule
\textbf{RevolveVLN (Ours)} & \textbf{38.0} & 12.9 & \textbf{37.7} & \textbf{44.3} & \textbf{43.0} & \textbf{26.5} & \textbf{4.1} & \textbf{46.7} & \textbf{33.7} \\
\bottomrule
\end{tabular*}
\par\vspace{2pt}
\begin{minipage}{\columnwidth}\fontsize{7}{8}\selectfont
\textbf{VP}, \textbf{PM}, \textbf{VS}, \textbf{SpR}, \textbf{ES}, and \textbf{SS} denote vertical perception, precise movement, viewpoint shifting, spatial relationship, environment state, and space structure, respectively. $\dagger$: trained without additional navigation data. The unmarked RevolveVLN version uses 1.10M additional training samples. Best results are in \textbf{bold}. Baseline results are sourced from the NavSpace benchmark~\cite{yang2026navspace}.
\end{minipage}
\endgroup
\end{table}

%% file: sections/05-conclusion.tex
\section{Conclusion}
\label{sec:conclusion}
We presented RevolveVLN, a monocular vision-language navigation framework built on \textit{Evolving Egocentric Memory}. Through \textit{Refresh}, \textit{Reinforce}, and \textit{Relate}, it continually updates historical geometry, enriches current observations, and retrieves pose-relative memory. Evaluations on R2R-CE, RxR-CE, and NavSpace demonstrate strong instruction following, remarkable data efficiency, and zero-shot spatial generalization, while real-world experiments validate deployment across complex indoor tasks. Looking ahead, we aim to enhance metric-level spatial precision for exact quantitative instructions in physical environments. Overall, our work underscores \textit{Evolving Egocentric Memory} as a promising path toward embodied spatial intelligence.

%% file: sections/appendix.tex
\appendix
\setlength{\parskip}{3pt plus 1pt minus 1pt}
\setcounter{table}{0}
\renewcommand{\thetable}{S\arabic{table}}
\renewcommand{\theHtable}{supp.\arabic{table}}
\setcounter{figure}{0}
\renewcommand{\thefigure}{S\arabic{figure}}
\renewcommand{\theHfigure}{supp.\arabic{figure}}

\section{Implementation and Training Details}
\label{app:training}

\subsection{Implementation Details}
\label{app:implementation}
\paragraph{Geometry and initialization.}
Each Relate attention head uses 48 channels for point geometry and 24 for camera geometry. 
All point coordinates and camera translations share a normalization factor given by the median valid depth in the current view.
Relate's query, key, and value projections are initialized from the corresponding native attention projections, and its mixing coefficient is initialized to $\alpha=0.05$.

\paragraph{Optimization.}
Relate's projections use the same learning rate as the language backbone; the remaining visual and geometric projection and fusion modules use $10^{-5}$.
The Reinforce gate parameters $\gamma_\ell$ and Relate gate parameter $\eta$ use $10^{-3}$ with zero weight decay.

\subsection{Training and Inference}
\label{app:training-inference}
\paragraph{Training.}
For each training sample, we reset $R^3$ and process the selected historical RGB frames (up to eight) followed by the current frame in chronological order.
No geometric state is shared across samples, even when they belong to the same trajectory.

\paragraph{Inference.}
At inference, $R^3$ retains its state within each episode and updates it once per new RGB observation, including observations received during action execution.
At each planning step, the visual input comprises the current frame and up to eight uniformly sampled historical frames.

\section{Additional Ablation Studies}
\label{app:ablations}

\subsection{Ablation Settings and Variants}
\label{app:ablation-definitions}

This section specifies the common settings and variant definitions for the ablations in the main paper and this supplement.
All variants follow the training setting without additional navigation data in Sec.~4.1 of the main paper: each is trained separately for one epoch on the same R2R-CE and RxR-CE expert demonstrations using the same training configuration.
Further training details are provided in Sec.~\ref{app:training}.
The ablations in Tables~3 and~4 of the main paper and Sec.~\ref{app:r2r-ablations} are evaluated on the R2R-CE validation-unseen split.
In Sec.~\ref{app:navspace-results}, we directly evaluate the same checkpoints corresponding to the two main-paper ablation tables on NavSpace, without additional training or NavSpace-specific fine-tuning.

Throughout the ablation studies, \emph{Full} denotes the complete model with Refresh, Reinforce, and Relate, corresponding to RevolveVLN$^{\dagger}$ trained without additional navigation data in Table~1 of the main paper.

\paragraph{Refresh.}
\emph{w/o Refresh} bypasses visual--geometric fusion before the LLM and feeds the unrefreshed visual tokens into the LLM, while still computing the current-view attention weights for reuse by Reinforce.
\emph{Per-frame geometry} retains cross-attention fusion but uses each view's initial geometric estimate from the time it was observed, instead of the geometric features updated at the current step.
\emph{Direct injection} keeps the geometry source fixed and directly injects geometric features into visual tokens before they enter the LLM, replacing cross-attention fusion.

\paragraph{Reinforce.}
\emph{w/o Reinforce} removes the additional geometry injections from the first three LLM decoder layers.
\emph{Independent attention} retains the geometry enhancement formulation but computes attention weights separately at each injection layer instead of reusing Refresh's current-view attention.
\emph{Direct injection} injects only the geometric features at each scale into the current-view hidden states in the first three LLM decoder layers; the \emph{non-residual} variant instead injects only geometric features aggregated through shared attention, without the direct geometry term.

\paragraph{Relate.}
\emph{w/o Relate} removes the geometric attention branch and restores the native attention output at the fourth LLM decoder layer.
\emph{Multi-layer placement} applies Relate at layers 4, 13, 19, 28, and 34 rather than only at layer 4.
\emph{Identity transforms} retains the attention branch, learnable projections, and mixing gate, while replacing the explicit 3D transformations in this branch with identity maps.
This replacement affects both attention matching and value aggregation and preserves the trainable parameter count.

\subsection{Fine-Grained Ablations on R2R-CE}
\label{app:r2r-ablations}

Table~\ref{tab:supp-fine-grained-ablation} further examines the fusion strategy in Refresh, geometry enhancement in Reinforce, and explicit 3D transformations in Relate on R2R-CE validation-unseen.

\begin{table}[!htbp]
\centering
\caption{\textbf{Fine-grained ablations on R2R-CE.}}
\label{tab:supp-fine-grained-ablation}
\small
\setlength{\tabcolsep}{3pt}
\begin{tabular*}{\columnwidth}{@{}l@{\extracolsep{\fill}}rrrr@{}}
\toprule
Variant & NE$\downarrow$ & OSR$\uparrow$ & SR$\uparrow$ & SPL$\uparrow$ \\
\midrule
\textbf{Full} & \textbf{4.25} & \textbf{67.86} & \textbf{64.22} & \textbf{59.35} \\
Refresh: direct injection & 4.79 & 64.27 & 60.79 & 56.27 \\
Reinforce: direct injection & 4.73 & 65.04 & 61.72 & 56.79 \\
Reinforce: non-residual & 4.44 & 65.04 & 58.35 & 53.38 \\
Relate: identity transforms & 4.44 & 65.85 & 61.83 & 57.28 \\
\bottomrule
\end{tabular*}
\end{table}

\paragraph{Cross-Attention Fusion vs. Direct Injection (Refresh).}
Replacing input cross-attention fusion with direct geometry injection reduces SR from \emph{Full}'s 64.22\% to 60.79\% and SPL from 59.35\% to 56.27\%, drops of 3.43 and 3.08~pts, respectively, while increasing NE from 4.25~m to 4.79~m.
Direct injection fuses geometric features at corresponding positions, whereas cross-attention allows each visual token to select relevant information from the current geometric context, enabling each view to extract geometry relevant to its visual content.
The improvements across these three metrics show that visually guided fusion makes more effective use of the same geometric context.

\paragraph{Residual Enhancement vs. Single-Path Injection (Reinforce).}
Relative to \emph{Full}, the \emph{direct injection} and \emph{non-residual} variants reduce SR by 2.50 and 5.87~pts and SPL by 2.56 and 5.97~pts, respectively.
The \emph{non-residual} variant achieves lower NE than \emph{direct injection} (4.44~m vs. 4.73~m), but has lower SR and SPL. The residual enhancement in \emph{Full} further reduces NE to 4.25~m and improves both SR and SPL.
Direct geometry preserves information at positions corresponding to the visual tokens, while shared attention adds relevant spatial content.
Combining both outperforms either signal alone across all three metrics, indicating that attention aggregation is more effective as an enhancement to the original geometry than as a complete replacement for geometry at corresponding positions.

\paragraph{Explicit 3D vs. Identity Transforms (Relate).}
Replacing explicit 3D transformations with identity maps reduces SR from \emph{Full}'s 64.22\% to 61.83\% and SPL from 59.35\% to 57.28\%, drops of 2.39 and 2.07~pts, respectively, while increasing NE from 4.25~m to 4.44~m.
The attention branch remains present, but 3D relative relations no longer explicitly participate in matching and content aggregation.
Although Refresh and Reinforce already introduce geometry into the visual representations, explicit 3D transformations further improve navigation, demonstrating the value of relative relations beyond geometry-enhanced content representations.
Relate directly incorporates relative positions and orientations in the current reference frame into cross-view associations, adding explicit spatial constraints to interactions between historical and current content.

\subsection{Complete Results on NavSpace}
\label{app:navspace-results}

To further disentangle the contributions of individual modules and key design choices to spatial reasoning under zero-shot transfer, Tables~\ref{tab:supp-navspace-components} and~\ref{tab:supp-navspace-designs} directly evaluate the variants from Table~3 (component ablations) and Table~4 (design-choice ablations) of the main paper on NavSpace across all six spatial task categories. VP, PM, VS, SpR, ES, and SS denote vertical perception, precise movement, viewpoint shifting, spatial relationship, environment state, and space structure, respectively. NE is reported in meters; the remaining metrics are percentages.

\begin{table}[!htbp]
\centering
\caption{\textbf{Component ablation on NavSpace.}}
\label{tab:supp-navspace-components}
\begingroup
\fontsize{8}{9}\selectfont
\setlength{\tabcolsep}{3.5pt}
\renewcommand{\arraystretch}{1.08}
\begin{tabular*}{\textwidth}{@{\extracolsep{\fill}}l*{6}{c}!{\hspace{8pt}}*{3}{c}@{}}
\toprule
& \multicolumn{6}{c}{Subtask SR (\%)} & \multicolumn{3}{c}{Macro average} \\
\cmidrule(lr){2-7}\cmidrule(l){8-10}
Variant & VP & PM & VS & SpR & ES & SS & NE$\downarrow$ & OSR$\uparrow$ & SR$\uparrow$ \\
\midrule
\textbf{Full} & \textbf{36.06} & \textbf{10.95} & 31.88 & 41.98 & \textbf{39.00} & \textbf{14.50} & \textbf{4.25} & \textbf{42.75} & \textbf{29.06} \\
w/o Refresh & 34.13 & 9.95 & 30.43 & 37.74 & 36.00 & 11.50 & \textbf{4.25} & 40.64 & 26.63 \\
w/o Reinforce & 27.88 & 8.96 & 26.57 & 26.42 & 30.50 & 9.00 & 4.63 & 32.62 & 21.55 \\
w/o Relate & 31.73 & \textbf{10.95} & \textbf{33.33} & \textbf{43.40} & 37.50 & 12.00 & 4.39 & 41.53 & 28.15 \\
\bottomrule
\end{tabular*}
\endgroup

\vspace{2pt}
\caption{\textbf{Design choice ablation on NavSpace.}}
\label{tab:supp-navspace-designs}
\begingroup
\fontsize{8}{9}\selectfont
\setlength{\tabcolsep}{3.5pt}
\renewcommand{\arraystretch}{1.08}
\begin{tabular*}{\textwidth}{@{\extracolsep{\fill}}l*{6}{c}!{\hspace{8pt}}*{3}{c}@{}}
\toprule
& \multicolumn{6}{c}{Subtask SR (\%)} & \multicolumn{3}{c}{Macro average} \\
\cmidrule(lr){2-7}\cmidrule(l){8-10}
Variant & VP & PM & VS & SpR & ES & SS & NE$\downarrow$ & OSR$\uparrow$ & SR$\uparrow$ \\
\midrule
\textbf{Full} & \textbf{36.06} & 10.95 & \textbf{31.88} & \textbf{41.98} & 39.00 & \textbf{14.50} & 4.25 & \textbf{42.75} & \textbf{29.06} \\
Refresh: per-frame geometry & 30.29 & \textbf{16.42} & 31.40 & 35.38 & \textbf{43.50} & 9.00 & 4.27 & 39.20 & 27.66 \\
Reinforce: independent attention & 32.21 & 9.45 & 28.02 & 39.62 & 35.00 & 10.50 & 4.36 & 40.25 & 25.80 \\
Relate: multi-layer placement & 30.29 & 11.44 & 29.95 & 37.74 & 37.00 & 10.50 & \textbf{4.16} & 38.60 & 26.15 \\
\bottomrule
\end{tabular*}
\endgroup
\end{table}

\noindent\textbf{Refresh.} Removing Refresh lowers SR in all six categories and reduces macro SR and OSR by 2.43 and 2.11~pts, respectively. The decreases concentrate on SpR (4.24~pts), ES (3.00~pts), and SS (3.00~pts), which require reassessing object relations, environment states, and scene structure as the viewpoint changes. This pattern shows that refreshing visual representations with the current geometric context helps maintain spatial consistency across time. Although macro NE remains 4.25~m, the lower OSR and SR further show that similar average endpoint distance does not ensure reliable goal-region entry or task completion.

\noindent\textbf{Reinforce.} Removing Reinforce causes the largest overall degradation: macro SR falls by 7.51~pts to 21.55\%, macro OSR falls by 10.13~pts to 32.62\%, and NE increases from 4.25~m to 4.63~m. The largest SR drops occur on SpR (15.56~pts), ES (8.50~pts), and VP (8.18~pts). These tasks require associating the current observation with geometric evidence accumulated from earlier viewpoints, showing that continued geometric enhancement in the early decoder layers is central to integrating evolving spatial memory.

\noindent\textbf{Relate.} Removing Relate reduces macro SR to 28.15\% and increases NE to 4.39~m. Its clearest effects appear on VP, SS, and ES, where SR decreases by 4.33, 2.50, and 1.50~pts, respectively. These categories more directly depend on organizing vertical direction, environment state, and global structure in the current egocentric reference frame. PM remains unchanged, while VS and SpR do not follow the same trend, indicating that Relate provides targeted benefits for coordinate-dependent 3D reasoning rather than uniform gains across all categories.

\noindent\textbf{Updated vs. Per-Frame Geometry (Refresh).} Replacing updated geometry with each frame's initial estimate lowers macro SR from 29.06\% to 27.66\% and macro OSR from 42.75\% to 39.20\%. The largest SR drops occur on SpR (6.60~pts), VP (5.77~pts), and SS (5.50~pts), showing that historical geometry must evolve with the current context to support cross-view relations and global structure. PM and ES do not follow the same trend, suggesting that Refresh primarily benefits relational and structural integration rather than every local navigation behavior equally.

\noindent\textbf{Shared vs. Independent Attention (Reinforce).} Computing attention independently at each injection layer lowers SR in all six categories, with macro SR decreasing to 25.80\% and macro OSR to 40.25\%. These results support reusing Refresh's visual--geometric correspondences beyond input fusion and suggest that shared attention provides more consistent geometric guidance across decoder layers than computing attention independently at each layer.

\noindent\textbf{Single- vs. Multi-Layer Placement (Relate).} Extending explicit 3D transformations from layer 4 to five layers lowers macro SR and OSR to 26.15\% and 38.60\%, respectively. Although NE improves from 4.25~m to 4.16~m, the lower endpoint error does not translate into more reliable goal-region entry or task completion. The focused placement at layer 4 therefore provides more reliable task success than repeatedly applying geometric transformations across multiple LLM decoder layers.

\section{Qualitative Analysis}
\label{app:qualitative}

\subsection{Mechanism Visualization}
\label{app:mechanism-visualization}

\subsubsection{Reinforce}
\label{app:reinforce-visualization}
\begin{figure}[!htbp]
  \centering
  \includegraphics[width=0.82\textwidth]{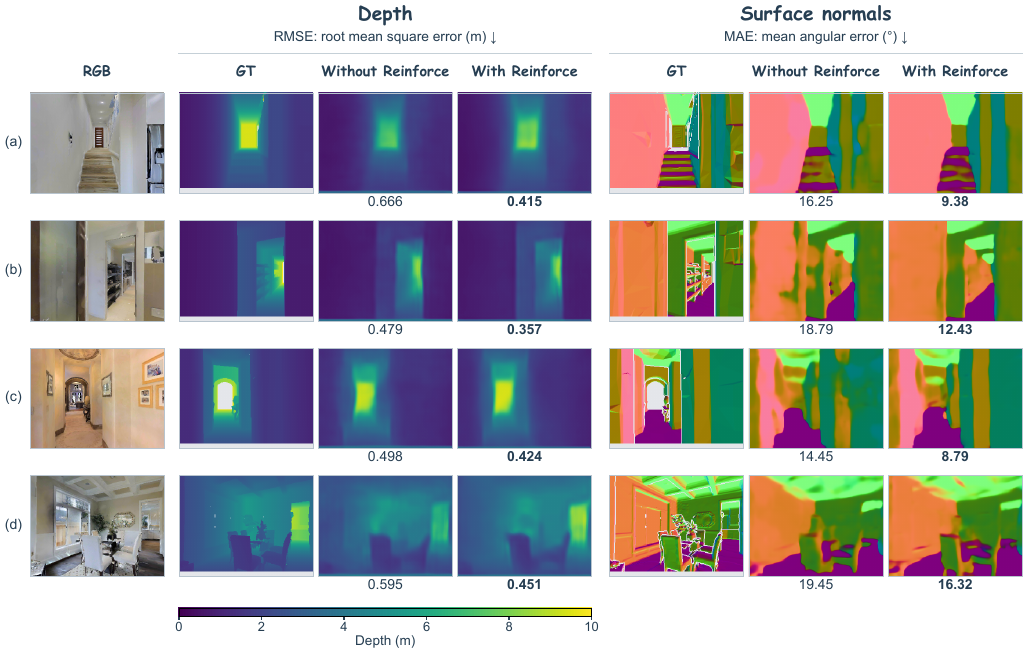}
  \caption{\textbf{Depth and surface-normal readouts with Reinforce.} \textbf{Left:} current RGB observations, one example per row. \textbf{Middle/right:} ground-truth depth/normals and predictions decoded from frozen representations without or with Reinforce. \textbf{Legend and metrics:} depth colors indicate distance ($0$--$10$\,m); normal colors encode orientation. Gray marks invalid GT regions. Values report per-image depth RMSE (m) and mean normal angular error (MAE, $^\circ$) over valid GT pixels.}
  \label{fig:reinforce_geometry}

\end{figure}

To examine whether Reinforce strengthens current-scene geometry encoding, we train depth and surface-normal readout heads under identical settings on frozen current-view navigation features to predict scene distances and surface orientations. As shown in Figure~\ref{fig:reinforce_geometry}, predictions without Reinforce underestimate distances to the far end of the corridor and the space beyond the doorway, while their normal maps mix the orientations of adjacent surfaces around doorways and wall corners. With Reinforce, the decoded depth more closely matches the ground-truth distance structure, and the normal maps better distinguish adjacent wall planes. In example (a), depth RMSE decreases from $0.666$ to $0.415$\,m, and mean angular error decreases from $16.25^\circ$ to $9.38^\circ$. These examples provide qualitative support for Reinforce's role in enriching navigation representations with scene geometry, as reflected in more accurate readout of distances and surface orientations from current-view features.

\subsubsection{Relate}
\label{app:relate-visualization}
\begin{figure}[!htbp]

\centering
\includegraphics[width=0.82\textwidth]{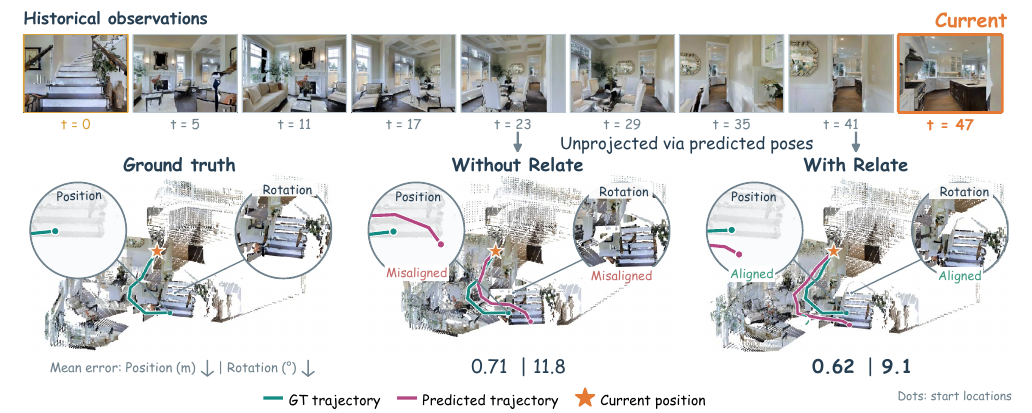}
\caption{\textbf{Historical camera pose readout and scene alignment with Relate.} \textbf{Top:} eight historical RGB observations and the current view. \textbf{Bottom:} scenes assembled using ground-truth poses or poses decoded without or with Relate, sharing RGB observations and GT depth used only for visualization. \textbf{Insets:} camera trajectories and cross-view structural alignment. \textbf{Legend and metrics:} teal/purple curves denote GT/predicted trajectories; dots mark starts and the star marks the current position. Values report mean camera position (m) and rotation ($^\circ$) errors.}
\label{fig:supp-relate-poses}
\end{figure}

To examine whether navigation representations encode 3D relationships between historical observations and the current view, we train pose readout heads under identical settings on frozen RGB-only navigation features to predict historical camera positions and orientations in the current egocentric frame. We combine these predictions with ground-truth depth to visualize the assembled scene. As shown in Figure~\ref{fig:supp-relate-poses}, after successive movements and viewpoint changes, historical poses predicted without Relate exhibit noticeable offsets, resulting in misaligned scene structures across views. With Relate, the predicted historical camera trajectory follows the ground-truth trajectory more closely, and scene structures align more consistently. In this sequence, mean position error decreases from 0.71~m to 0.62~m and mean rotation error from $11.8^\circ$ to $9.1^\circ$. These results indicate that Relate helps navigation representations retain relative spatial relationships between historical observations and the current view, enabling more consistent organization of cross-view information in the current reference frame.

\begin{samepage}
\subsection{Navigation Examples}
\label{app:simulation-cases}
Figures~\ref{fig:app-vln-cases} and~\ref{fig:app-navspace-cases} illustrate qualitative trajectories across R2R-CE, RxR-CE, and zero-shot NavSpace. On R2R-CE and RxR-CE, the agent tracks successive landmarks across rooms, such as passing the kitchen bar before turning at the stairs or circling a bed to reach a far nightstand. On NavSpace (Fig.~\ref{fig:app-navspace-cases}), the six examples span all six spatial reasoning categories. Several cases directly test coherent 3D scene structure across large heading rotations and viewpoint shifts: the agent adopts the TV's perspective to reach the dining table (Row~3), branches upstairs after spotting a potted plant (Row~5), and executes a $180^\circ$ loop around a bed to stop at the opposite lamp (Row~6). Together, these cases demonstrate consistent spatial tracking across room transitions, turns, and conditional instructions.
\par\end{samepage}

\begingroup
\makeatletter
\setlength{\@fptop}{0pt}
\setlength{\@dblfptop}{0pt}
\makeatother
\begin{figure}[!htbp]
\centering
\includegraphics[width=0.995\textwidth]{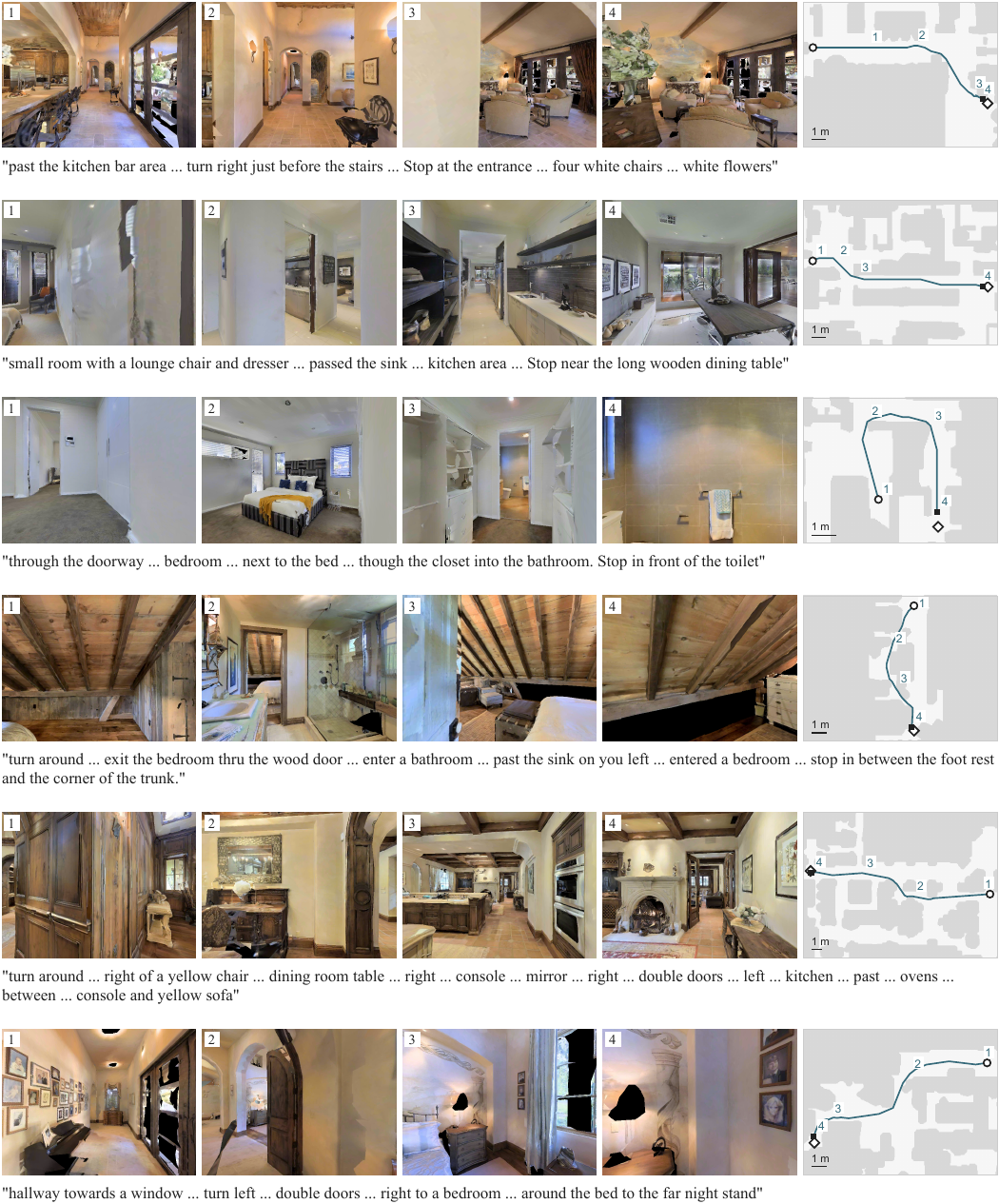}
\caption{\textbf{Navigation examples on R2R-CE and RxR-CE.} The first three rows show R2R-CE and the last three RxR-CE. Each row pairs an instruction excerpt with four observations and the full-model trajectory. Numbers locate the observations on the map. Blue lines show the path; circles, diamonds and squares mark the start, goal and stop. Map scales are shown separately.}
\label{fig:app-vln-cases}
\end{figure}

\begin{figure}[!htbp]
\centering
\includegraphics[width=0.995\textwidth]{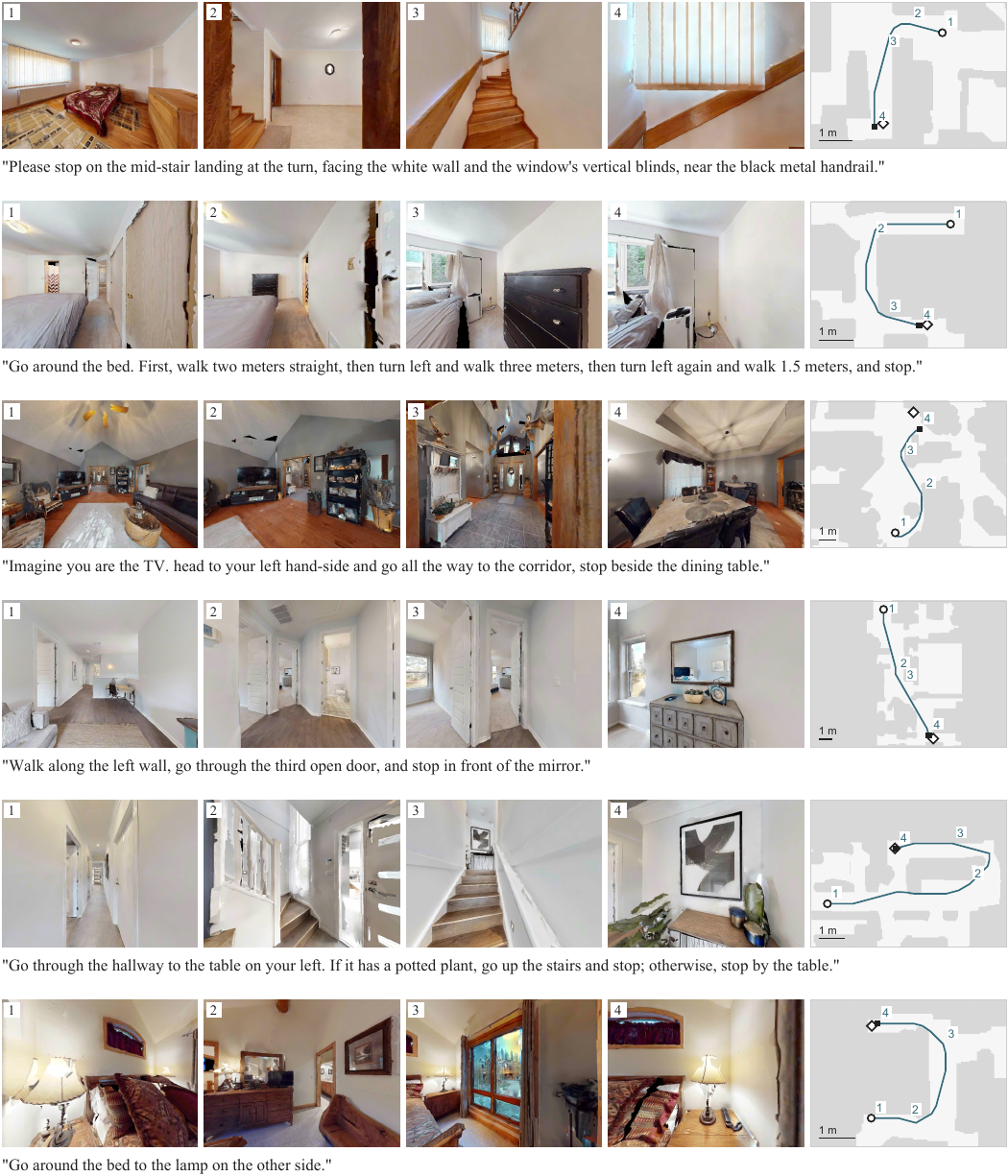}
\caption{\textbf{Navigation examples on NavSpace.} From top to bottom: vertical perception, precise movement, viewpoint shifting, spatial relationship, environment state, and space structure. Four numbered observations accompany each instruction and trajectory. Map symbols follow Fig.~\ref{fig:app-vln-cases}.}
\label{fig:app-navspace-cases}
\end{figure}

\endgroup

\section{Real-World Deployment and Evaluation}
\label{app:real-world}

\subsection{Continuous Trajectory Generation}
\label{app:action-to-trajectory}
Starting from the odometry pose associated with the input image, we integrate each predicted action chunk into a planar path using 0.25\,m forward steps and $15^\circ$ turns.
Adjacent identical actions are merged, while the endpoints of forward segments are retained as waypoints.
When a new prediction becomes available, we use the current pose to remove the portion of the predicted path already traversed by the robot, keeping the remaining waypoints fixed in the odometry frame.
We then connect the current pose and the remaining waypoints with cubic Hermite curves.
A low-level controller tracks translational paths and executes pure rotations in place.

\subsection{Computational Resources}
\label{app:gpu-memory}
We deploy the version of RevolveVLN trained without additional data on a single NVIDIA RTX 5090 D v2 GPU. The maximum observed GPU memory usage of the model process is 15.79\,GiB, measured at a sampling rate of 1\,Hz.

\subsection{Real-World Navigation Examples}
\label{app:robot-examples}
\begin{figure}[!htbp]
  \centering
  \includegraphics[width=0.71\textwidth]{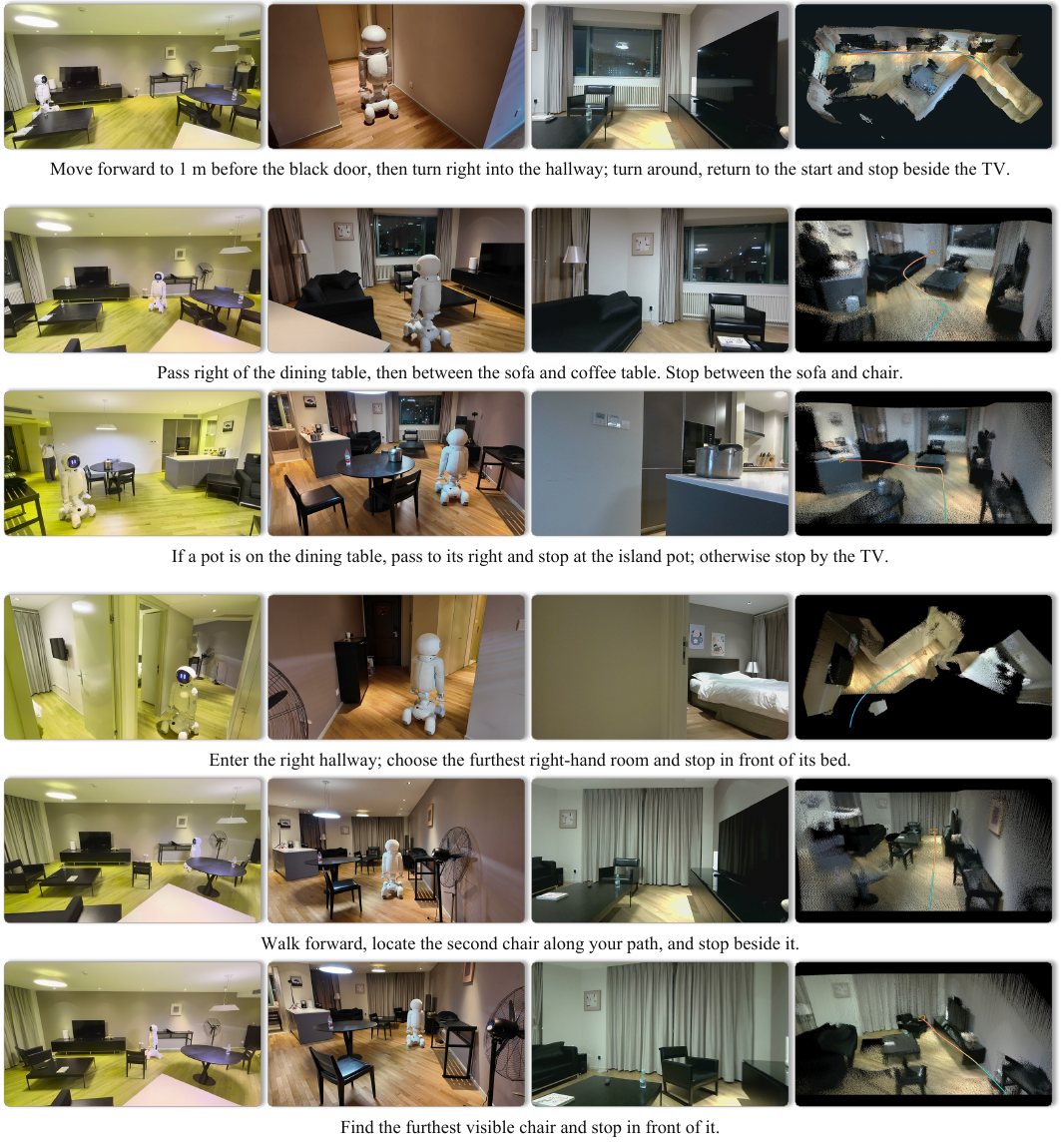}
  \caption{\textbf{Real-world navigation examples.} Columns show fixed-camera, moving third-person, egocentric, and $R^3$ memory views.}
  \label{fig:supp-robot-examples}

\vspace{6pt}
\includegraphics[width=0.71\textwidth]{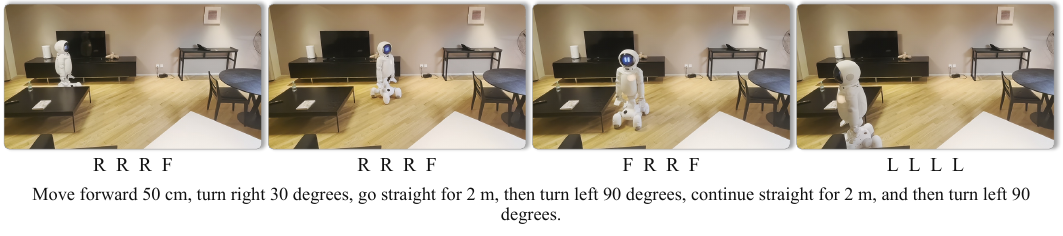}
  \caption{\textbf{Failure case.} Four chronological frames with approximately aligned predicted actions (F: forward; L/R: left/right).}
  \label{fig:supp-robot-failure}
\end{figure}

Figure~\ref{fig:supp-robot-examples} presents six real-world navigation examples using fixed-camera, moving third-person, egocentric, and $R^3$ memory visualizations. The tasks cover route revisiting, relational and conditional navigation, and ordinal and distance-based goal selection.

\subsection{Supplementary Video}
\label{app:robot-video}
The supplementary videos provide continuous recordings with instructions, first-person and external views, and geometric reconstructions where available.

\subsection{Limitations}
\label{app:robot-failures}
Figure~\ref{fig:supp-robot-failure} shows a failure on an instruction with explicit distances and turning angles, corresponding to the Precise Movement category in NavSpace.
The model initially predicts right turns instead of the requested forward movement and later fails to recover when blocked near the sofa, leaving the route incomplete.
Our deployed model has not been specifically fine-tuned on metric navigation instructions, which may contribute to its difficulty in following such ordered action sequences.